\documentclass[runningheads]{llncs}

\usepackage{eccv}

\usepackage{eccvabbrv}

\usepackage{graphicx}
\usepackage{booktabs}
\usepackage{pifont} 

\usepackage{multirow} 
\usepackage[table]{xcolor}
\usepackage{enumitem}

\newcommand{\cmark}{\checkmark}%
\newcommand{\xmark}{$\times$}%

\usepackage[accsupp]{axessibility}  

\usepackage{hyperref}

\usepackage{orcidlink}

\begin{document}

\title{PnP-Transformer: A Neural Pose Solver for Zero-Shot 6D Object Pose Estimation} 

\titlerunning{PnP-Transformer: A Neural Pose Solver for 6D Object Pose Estimation}

\author{Emirkan Burak Yilmaz \and
Yakup Genc}

\authorrunning{E.~B.~Yilmaz and Y.~Genc}

\institute{Computer Engineering Dept., Gebze Technical University, Kocaeli, Türkiye\\
\email{\{emirkan.yilmaz, yakup.genc\}@gtu.edu.tr}}

\maketitle

\begin{abstract}
We present PnP-Transformer, a neural pose solver designed as an iterative-free, differentiable alternative to classical PnP-RANSAC algorithms. PnP-Transformer regresses the 6D object pose by processing predicted semi-dense 2D--3D correspondence fields alongside confidence scores, while mapping geometric inputs into a continuous latent space. To evaluate its practical utility, we integrate PnP-Transformer into a zero-shot 6D pose estimation pipeline, where the correspondence module and pose solver are trained separately. PnP-Transformer achieves a $41.2\%$ mean AR score with low latency ($0.46$\,s/obj) on seven core BOP benchmark datasets, surpassing all prior baselines (such as RayPose with $39.1\%$ AR) but one (Co-op with $58.4\%$ AR). When the pose solver is replaced with EPnP-RANSAC (similar to Co-op) while keeping the correspondence module unchanged, the RANSAC alternative outperforms PnP-Transformer on six of the seven datasets by reaching $52.2\%$ mean AR at a cost of an additional $60$\,ms per object, revealing an accuracy-latency trade-off. Code and pretrained models are available: \url{https://github.com/ebylmz/pnp-transformer}

\end{abstract}

\section{Introduction}
\label{sec:intro}

Estimating the 6-Degree-of-Freedom (6DoF) pose of objects from monocular images is a fundamental problem in spatial AI and computer vision. In robotics, accurate prediction of the 6D pose allows embodied agents to grasp and manipulate objects precisely, facilitating automated warehouse operations and complex assembly tasks. In augmented reality, establishing precise spatial alignment between physical objects and virtual overlays is essential for seamless, immersive user experiences. 

Traditionally, pose estimation pipelines have relied on a two-stage decoupled architecture: first, predicting intermediate geometric representations such as 3D scene coordinates; second, solving for the rigid transformation using a variant of PnP\cite{gao2003complete} algorithm wrapped inside the RANSAC\cite{fischler1981random} solver. Although those solvers provide robust geometric model fitting, their non-differentiable nature does not allow the gradients to flow backward from the final 3D pose loss to the feature extraction. Consequently, neural networks in these frameworks are forced to rely on surrogate coordinate-regression or reprojection losses during training, rather than optimizing the primary rotation and translation metrics. However, optimizing intermediate representations independently of the downstream 3D spatial alignment task leads to sub-optimal performance, as small errors in predicted 2D keypoints can result in completely different poses during geometric fitting. To bridge this gap, recent works have proposed differentiable geometric solvers that integrate pose optimization directly into end-to-end training computation graphs. Prominent paradigms include probabilistic hypothesis selection\cite{brachmann2017dsac}, implicit differentiation via the Implicit Function Theorem\cite{chen2020end}, continuous probability density modeling on $\mathrm{SE}(3)$\cite{chen2022epro}, and linearized covariance supervision\cite{liu2023linear}. Although these methods restore gradient flow, they introduce substantial computational and numerical challenges. Optimization-based differentiable layers depend heavily on iterative solver loops (e.g., Gauss-Newton or Levenberg-Marquardt), matrix inversions, or expensive Monte Carlo sampling procedures. These operations introduce significant memory overhead, slow down training convergence, and remain prone to gradient divergence under ill-conditioned Hessians caused by noisy, degenerate, or ambiguous point distributions. Alternatively, direct regression architectures\cite{wang2021gdr, wen2024foundationpose} bypass explicit perspective solvers entirely by mapping learned dense representations directly to rotation and translation. Although these feed-forward networks achieve impressive speed, they discard explicit perspective projection constraints. Consequently, unconstrained regression layers often exhibit poor outlier resilience under severe occlusions, spatial domain shifts, and complex object geometries.

To overcome this divide, we introduce PnP-Transformer, a transformer-based geometric solver that maps 2D--3D correspondence fields directly to 6D object poses. Rather than treating PnP as an unlearnable, external post-processing step or discarding explicit geometric constraints entirely, PnP-Transformer embeds spatial relationships into self-attention blocks while learning a structured latent space. To demonstrate its practical utility, we integrate PnP-Transformer into a zero-shot 6D pose estimation pipeline for unseen objects, operating on a single RGB scene image and a target 3D CAD model. Given a scene image, our pipeline extracts semi-dense 2D--3D correspondence fields and confidence scores between the query image and a sparse set of rendered template views. The correspondence field from the highest-confidence template is passed to PnP-Transformer to regress the relative transformation between the query and template frames, which is composed with the reference template pose to recover the final 6D object pose. When evaluated across seven core BOP benchmark datasets, our pipeline operating with PnP-Transformer demonstrates competitive performance at low latency. Furthermore, substituting EPnP-RANSAC for our neural solver, while keeping the correspondence module fixed, yields higher overall accuracy. In summary, our main contributions are as follows:
\begin{enumerate}
    \item We introduce PnP-Transformer, a neural pose solver designed as a differentiable, iterative-free alternative to classical PnP-RANSAC algorithms.
    \item We propose a joint-embedding architecture that maps 2D--3D correspondence fields and direct 6D target poses into a shared latent space.
    \item We integrate PnP-Transformer within a zero-shot 6D pose estimation pipeline and evaluate across seven core BOP benchmark datasets. Our model achieves $41.2\%$ mean AR score at $0.46$\,s/obj, outperforming prior baselines such as RayPose ($39.1\%$) and FoundPose ($37.3\%$), while trailing only Co-op ($58.4\%$). 
    \item We perform a controlled evaluation comparing our neural pose solver against an EPnP-RANSAC alternative by swapping only the solver module while holding the correspondence network fixed. This reveals a direct accuracy-for-latency trade-off: the EPnP-RANSAC variant achieves $52.2\%$ mean AR at $0.52$\,s/obj, outperforming PnP-Transformer on six of the seven datasets by an average of $+11.0\%$ AR at the cost of an additional $60$\,ms per object.
\end{enumerate}

\section{Related Work}
\label{sec:rel_work}

\subsubsection{Differentiable Geometric Solvers.} 

Geometry-driven pose estimation methods establish 2D--3D correspondences from RGB images and recover the object pose using a PnP algorithm \cite{gao2003complete} combined with RANSAC \cite{fischler1981random}. However, non-differentiable operations in standard solvers (such as discrete inlier counting and hard $\arg\max$ selection) prevent backpropagation during end-to-end training. To enable gradient flow, recent works integrate geometric optimization into neural computational graphs. Differentiable RANSAC (DSAC) \cite{brachmann2017dsac} replaces hard hypothesis selection with a continuous probabilistic expectation over sampled subsets, enabling scene coordinate regression driven by pose loss. To avoid sampling loops, BPnP \cite{chen2020end} formulates PnP as a declarative layer, leveraging the Implicit Function Theorem (IFT) on the stationary optimality conditions of the non-linear least-squares objective. BlindPnP \cite{campbell2020solving} incorporates Sinkhorn optimal transport layers to estimate camera pose without requiring explicit prior matches. To address gradient instabilities caused by multimodal ambiguities on symmetric objects, EPro-PnP \cite{chen2022epro} models PnP output as a continuous probability density over $\mathrm{SE}(3)$, optimizing Kullback-Leibler (KL) divergence via Monte Carlo sampling. Meanwhile, the Linear-Covariance Loss \cite{liu2023linear} linearizes the PnP solver around ground-truth poses to supervise pose distribution covariance directly, eliminating gradient confusion caused by PnP averaging.

\subsubsection{Direct Pose Regressors.} An alternative paradigm bypasses geometric solvers entirely by using deep architectures to directly map visual features to 6D pose. To avoid topological discontinuities during direct rotation regression (such as gimbal lock or double-cover ambiguities in quaternions), modern networks adopt continuous 6D rotation parameterizations\cite{zhou2019continuity}, which stabilize gradient updates during training. Direct regressors like GDR-Net\cite{wang2021gdr} combine dense 2D--3D coordinate maps with surface region maps through patch-based convolutional modules (Patch-PnP) to predict rotation matrices and translation vectors. FoundationPose\cite{wen2024foundationpose} proposed a transformer-based iterative refinement network and contrastive pose scoring modules, enabling zero-shot generalization across novel objects. 

\subsubsection{Unseen Object Pose Estimation.} Model-based approaches for novel object pose estimation \cite{moon2025co, caraffa2025accurate, huang2024matchu, moon2024genflow, ornek2024foundpose, nguyen2024gigapose, labbe2022megapose} can be broadly categorized into feature matching and template matching \cite{liu2024deep}. Feature-matching methods extract local descriptors from the input image, match them to features of the 3D model to establish 2D--3D correspondences, and recover the pose using the PnP~\cite{lepetit2009ep} algorithm. Notable examples include MatchU~\cite{huang2024matchu}, which combines 2D texture and 3D geometric cues using rotation-invariant descriptors and an attention-based fusion mechanism. Template-matching methods\cite{labbe2022megapose, nguyen2024gigapose, moon2024genflow, ornek2024foundpose, moon2025co} instead render synthetic views of the 3D model from multiple poses and estimate the pose by retrieving the closest matching template and aligning it with the observed image. MegaPose~\cite{labbe2022megapose} exemplifies this strategy by generating millions of synthetic images to train a deep network for large-scale pose retrieval. GigaPose~\cite{nguyen2024gigapose} improves efficiency by reducing the number of templates considered by focusing only on the out-of-plane rotations and learning rotation-invariant features. 

\subsubsection{Vision Foundation Models.} Recent advances in self-supervised vision foundation models provide strong visual representations for unseen object pose estimation. Models such as DINO~\cite{oquab2023dinov2, simeoni2025dinov3} and CroCo~\cite{weinzaepfel2023croco} produce stable, transferable patch descriptors that generalize across categories, viewpoints, and imaging conditions. FoundPose~\cite{ornek2024foundpose} demonstrated that patch descriptors extracted from different DINOv2 layers encode varying levels of positional and semantic cues, enabling effective correspondence-based pose estimation. Similarly, Co-op~\cite{moon2025co} leverages CroCov2 to extract patch-level features and compute semi-dense correspondences between the input image and pre-rendered templates. Together, these developments highlight the strength of foundation-model features as powerful appearance priors.

\section{Method}
\label{sec:method}

\subsection{Overview}
\begin{figure}[t]
    \centering
    \includegraphics[width=\linewidth]{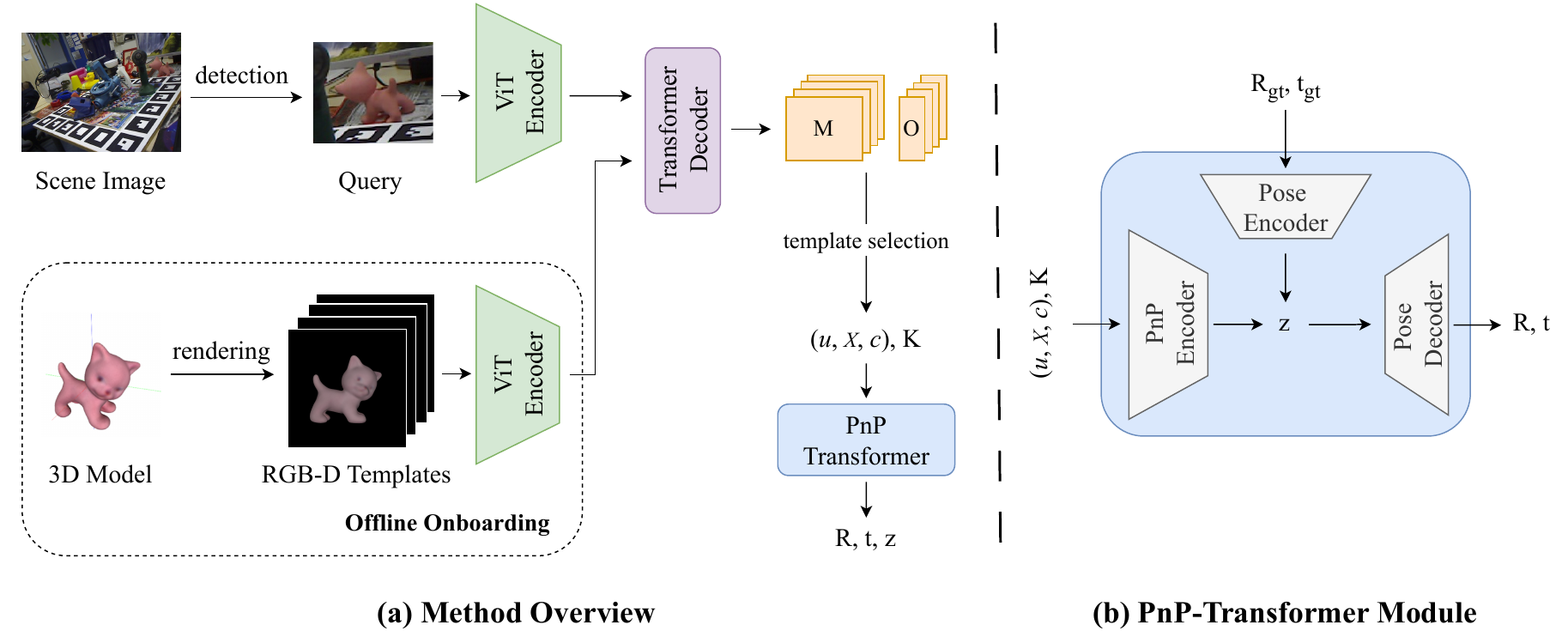}
    \caption{\textbf{Proposed architecture.} 
    (a) \textit{Method Overview:} During an offline onboarding stage, RGB-D templates are rendered from the 3D CAD model and processed through a ViT encoder to acquire the template features. At inference, a query image crop is similarly processed using the same encoder to obtain query features. A transformer decoder and following heads then predict the patch-level match matrix $\mathbf{M}$ and displacement offset $\mathbf{O}$ across all query-template pairs. The template with the highest overall match confidence is selected, and its corresponding 2D--3D correspondence field with associated point-wise confidences $(u, X, c)$ is fed into PnP-Transformer to regress the relative transformation $\Delta = (\mathbf{R}, \mathbf{t}) \in \mathrm{SE}(3)$ while producing the latent representation $\mathbf{z}$ of the recovered transformation.
    (b) \textit{PnP-Transformer Module:} PnP-Transformer consists of three primary sub-modules: \textit{PnP-Encoder} that processes input correspondences into a geometric latent vector $\mathbf{z}$, \textit{Pose-Encoder} used during training to map ground-truth target poses $(\mathbf{R}_{\text{gt}}, \mathbf{t}_{\text{gt}})$ into target embeddings $\bar{\mathbf{z}}$, and \textit{Pose-Decoder} that regresses final relative 6D pose parameters from $\mathbf{z}$.
    }
    \label{fig:method_overview}
\end{figure}

An overview of our pipeline is presented in Fig.~\ref{fig:method_overview}. Given a query RGB image $I^Q$, and the CAD model of a target object, our goal is to estimate its 6D pose without object-specific fine-tuning. We render a set of $K$ template views $\{ I^T_k \}_{k=1}^K$ from the CAD model. For every template, we predict 2D--3D semi-dense correspondences between $I^Q$ and $I^T_k$ and select the best template based on the match confidences. The correspondence field from the selected template is then processed by PnP-Transformer to recover the relative transformation $\Delta = (\mathbf{R}, \mathbf{t}) \in \mathrm{SE}(3)$, while encoding the geometric input into a latent representation $\mathbf{z}$.

\subsection{Object Onboarding}
\label{sec:onboarding}

For each target object, we construct a set of RGB-D templates by rendering its CAD model from multiple viewpoints. The camera poses are obtained by subdividing the triangular faces of an icosahedron, yielding approximately uniform coverage over viewing directions~\cite{nguyen2023cnos}. Similar to GigaPose~\cite{nguyen2024gigapose}, we limit the template set to out-of-plane rotations, reducing the computational cost of the template-matching approach. Specifically, applying a single level of subdivision on the icosahedron yields an icosphere with $K = 42$ discrete viewpoints. With the object centered at the world origin, the camera is positioned at each viewpoint to render the template RGB images $\{ I^T_k \}_{k=1}^K$ along with their corresponding depth maps and camera matrices. We pre-extract the patch-level features by passing each rendered template image through a Vision Transformer (ViT)\cite{dosovitskiy2020image} encoder, and store the template token features $\{ \mathbf{F}^T_k \}_{k=1}^K \in \mathbb{R}^{N \times D}$. This object onboarding stage is performed offline once per object. The stored features and depth maps are subsequently retrieved during online inference.

\subsection{Estimating 2D--3D Correspondences}

Our online pipeline begins by establishing semi-dense patch-level 2D--3D correspondences between the query image and the onboarded object templates. As illustrated in Fig.~\ref{fig:method_overview}, our matching module features a shared ViT encoder~\cite{dosovitskiy2020image} for feature extraction, paired with a transformer decoder~\cite{vaswani2017attention} utilizing cross-attention blocks to blend and interpret the query and template representations. We use an off-the-shelf object detector CNOS~\cite{nguyen2023cnos} to predict the bounding box of the target object within the scene image. The region of interest is then cropped and resized to a standard resolution of $H \times W$ to form the query image $I^Q$. The query image is then processed by the ViT encoder to extract the patch-level feature map $\mathbf{F}^Q \in \mathbb{R}^{N \times D}$, where $N= \frac{H}{16} \times \frac{W}{16}$ denotes the total number of patches, dividing by 16 accounts for the patchify operation in ViT, and $D$ is the feature dimension. Similar to patch classification and offset regression formula introduced in Co-op\cite{moon2025co}, the decoder and following heads process $\mathbf{F}^Q$ and $\mathbf{F}^T$, and produce the match matrix $\mathbf{M} \in \mathbb{R}^{N \times (N+1)}$ and offset $\mathbf{O} \in \mathbb{R}^{N \times 2}$ to refine the predicted patch matches within patch boundaries. Each row of $\mathbf{M}$ defines a probability distribution over $N$ template patches and an additional no-match class (as index $N$) to model background/occlusion in the query image.

To establish 2D--3D correspondences, each query patch center is first assigned to its most likely template patch counterpart via a discrete $\operatorname{argmax}$ selection over the row-wise match probabilities in $\mathbf{M}$. To achieve sub-patch precision, this discrete assignment is continuously refined by adding the predicted coordinate offset $\mathbf{O}$ to the selected template patch center. The refined 2D template coordinates are subsequently backprojected into the template camera frame using the pre-rendered template depth map, yielding the absolute 3D coordinate for each query patch center. We define a confidence value $c_i \in [0, 1]$ for each correspondence as:
\begin{equation}
c_i = 1 - \mathbf{M}[i, N],
\end{equation}
representing the complement of the predicted no-match probability. This correspondence estimation process is executed across all $K$ templates. Each template is scored by summing the matching probabilities of its selected patches while excluding the no-match predictions. The template yielding the highest score is selected as the optimal view, and its associated 2D--3D correspondences are forwarded to PnP-Transformer to recover the relative template-to-query camera transformation.

\subsection{Pose Recovery via PnP-Transformer}

Adopting the formulation of the Perspective-$n$-Point (PnP) problem~\cite{gao2003complete}, we propose a joint-embedding architecture, termed \textbf{PnP-Transformer}, to regress the relative rigid transformation $\Delta = (\mathbf{R}, \mathbf{t}) \in \mathrm{SE}(3)$ from 2D--3D correspondence fields while learning a structured latent space. As illustrated in Fig.~\ref{fig:method_overview}(b), the architecture comprises three core modules: (1) \textbf{PnP-Encoder} that maps predicted correspondences to a latent vector $\mathbf{z} \in \mathbb{R}^{256}$, (2) \textbf{Pose-Decoder} that regresses the relative transformation $\Delta$ from $\mathbf{z}$, and (3) \textbf{Pose-Encoder} that maps ground-truth transformations into a target latent vector $\bar{\mathbf{z}} \in \mathbb{R}^{256}$.

\subsubsection{PnP-Encoder.} The PnP-Encoder module is a transformer-based network that processes semi-dense correspondence fields into a compact latent vector $\mathbf{z} \in \mathbb{R}^{256}$ representing the underlying transformation. Given $N$ predicted correspondences, the input consists of 2D query pixel coordinates $\mathbf{u}_i \in \mathbb{R}^{2}$, corresponding 3D template points $\mathbf{X}_i \in \mathbb{R}^{3}$, match confidence scores $c_i \in [0, 1]$, and the query camera intrinsics $\mathbf{K}$. To ensure invariance to varying camera parameters, we first transform all 2D pixel coordinates $\mathbf{u}_i = [u_i, v_i]^\top$ into normalized image space:
\begin{equation}
\mathbf{\tilde{u}}_i = \mathbf{K}^{-1} [u_i, v_i, 1]^\top.
\end{equation}
For each correspondence $i$, we construct a 6-dimensional input vector $\mathbf{v}_i = [\mathbf{\tilde{u}}_i^\top, \mathbf{X}_i^\top, c_i]^\top$. A linear layer projects each $\mathbf{v}_i$ into a $D$-dimensional token. A learnable class token is appended to the sequence of $N$ correspondence tokens, yielding an input feature matrix $\mathbf{Z}_{\text{in}} \in \mathbb{R}^{(N + 1) \times D}$. This sequence is then processed through a stack of transformer self-attention layers, where global spatial geometry and relative spatial constraints are aggregated across tokens. Finally, the processed class token is passed through an MLP to regress the latent embedding $\mathbf{z} \in \mathbb{R}^{256}$.

\subsubsection{Pose-Decoder.} The Pose-Decoder module regresses the relative template-to-query transformation $(\mathbf{R}, \mathbf{t})$ from the latent vector $\mathbf{z}$. It is parameterized as a lightweight MLP comprising two shared layers followed by independent prediction heads for rotation and translation. The rotation head predicts an unconstrained continuous 6D representation $[\mathbf{a}_1 \mid \mathbf{a}_2] \in \mathbb{R}^{3 \times 2}$~\cite{zhou2019continuity}, which is converted to a valid rotation matrix $\mathbf{R} \in \mathrm{SO}(3)$ via Gram-Schmidt orthogonalization. The translation head directly predicts the 3D translation vector $\mathbf{t} \in \mathbb{R}^3$ in camera space.

\subsubsection{Pose-Encoder.} The Pose-Encoder module acts as an auxiliary regularization module during training to structure the latent embedding space. It takes the ground-truth transformation $(\bar{\mathbf{R}}, \bar{\mathbf{t}})$ as input in 9D representation \cite{zhou2019continuity}, and projects it through an MLP to output the target latent embedding $\bar{\mathbf{z}} \in \mathbb{R}^{256}$. By enforcing an alignment between the predicted embedding $\mathbf{z}$ by PnP-Encoder and the target embedding $\bar{\mathbf{z}}$ during training, the model learns a topologically consistent representation of the $\mathrm{SE}(3)$ manifold.

\subsection{Training}
\label{sec:training_loss_functions}

\subsubsection{MegaPose Dataset.} We trained our models on the MegaPose-GSO dataset\cite{labbe2022megapose}, which provides photorealistic multi-object scene renderings of assets with object pose annotations, depth maps, and object masks. To construct our training query-template pairs, we randomly sample up to 4 objects per scene, enforcing a visibility threshold where at least 10\% of the object's surface must be visible, similar to the protocol of the BOP datasets\cite{nguyen2025bop}. Each selected object instance is cropped based on its region-of-interest bounding box and resized to a standard input resolution of $224 \times 224$ pixels to form the query image, which is paired with its geometrically closest template image with the same image resolution based on out-of-plane rotation. Ground-truth 2D--3D patch correspondences are constructed by unprojecting the query patch centers into the query camera frame using the ground-truth depth. These 3D points are then mapped to the template camera frame via ground-truth relative transformation, and projected onto the template image plane to identify the nearest template patch index and calculate the sub-patch offset. We apply a 5\% relative depth threshold during the projection step. Query points that are projected outside the template mask or exceed the depth error tolerance are assigned to the no-match class, and their corresponding ground-truth offset vectors are set to zero.

\subsubsection{Training Schedules.} We optimize the correspondence decoder and the PnP-Transformer modules separately. Although PnP-Transformer itself is differentiable, the discrete $\operatorname{argmax}$ operation used for patch selection during correspondence estimation breaks the gradient flow back into the matching network, preventing joint end-to-end training. Consequently, both modules are trained separately using distinct objective functions: the correspondence decoder is optimized via match loss, whereas PnP-Transformer is optimized using pose loss.

\subsubsection{Match loss.} To optimize the correspondence decoder and its heads, we minimize a joint objective combining a classification term for patch matching and a regression term for continuous sub-patch displacements\cite{moon2025co}. The patch classification loss $\mathcal{L}_{\text{cls}}$ is formulated as a standard cross-entropy loss over the match matrix $\mathbf{M}$ against the generated ground-truth patch indices. For the displacement vectors, we minimize the $\ell_1$ distance between the predicted offsets $\mathbf{O}$ and the ground-truth offsets, denoted as $\mathcal{L}_{\text{off}}$. The total match loss is defined as:
\begin{equation}
\mathcal{L}_{\text{match}} = \mathcal{L}_{\text{cls}} + \lambda_{\text{off}} \mathcal{L}_{\text{off}},
\end{equation}
where $\lambda_{\text{off}}$ is a hyperparameter balancing sub-patch offset regression loss relative to patch classification loss.

\subsubsection{Pose loss.} To supervise the pose regression components, we use the geodesic distance for rotation loss $\mathcal{L}_{\text{rot}}$, and the $\ell_2$ distance for translation loss $\mathcal{L}_{\text{trans}}$. Furthermore, to align the latent representations produced by PnP-Encoder and Pose-Encoder, we apply an $\ell_2$ consistency loss $\mathcal{L}_{\text{lat}}$ between the predicted embedding $\mathbf{z}$ and the ground-truth target embedding $\bar{\mathbf{z}}$, making the overall pose loss as:
\begin{equation}
\mathcal{L}_{\text{pose}} = \mathcal{L}_{\text{rot}} + \lambda_{\text{trans}} \mathcal{L}_{\text{trans}} + \lambda_{\text{lat}} \mathcal{L}_{\text{lat}},
\end{equation}
where $\lambda_{\text{trans}}$ and $\lambda_{\text{lat}}$ are the hyperparameters balancing the relative contributions of translation regression and latent manifold alignment, respectively.

\section{Experiments}
\label{sec:experiments}

\subsection{Experimental Setup}

\subsubsection{Implementation Details.} Our correspondence estimation network builds upon CroCo\cite{weinzaepfel2023croco}, a vision foundation model pretrained specifically for 3D vision tasks. We initialize our shared ViT encoder and transformer decoder with the pretrained weights of CroCoV2 to leverage its large-scale pretraining. During training, the ViT encoder is kept frozen, and we optimize only the weights of the transformer decoder and its prediction heads, which are initialized from scratch. Specifically, we employ the ViT-Large variant (24 layers, feature dimension $D=1024$) for the encoder and a ViT-Base architecture (12 layers, feature dimension 768) for the decoder. Each decoder prediction head is implemented as a single linear layer. For PnP-Transformer, PnP-Encoder incorporates 4 self-attention layers with a feature dimension of 1024, compressing the correspondence field into a 256-dimensional latent space. Both Pose-Encoder and Pose-Decoder are parameterized as 2-layer MLPs with a hidden dimension of 512. 

All modules are trained using the AdamW optimizer on a single NVIDIA GeForce RTX 4090 GPU, with a cosine learning rate schedule decaying from a peak of $2 \times 10^{-5}$ down to $1 \times 10^{-6}$. The correspondence decoder and prediction heads are trained with a batch size of 32 for 500k steps ($\sim$3 days) using the matching loss with offset weight $\lambda_{\text{off}} = 2.0$. PnP-Transformer is optimized in a two-stage regime: it is first pretrained on ground-truth 2D--3D correspondences with a batch size of 128 for 50k steps ($\sim$5 hours), followed by fine-tuning on noisy predicted correspondences from the decoder for 250k steps ($\sim$3 days) using pose loss with weights $\lambda_{\text{trans}} = 10.0$ and $\lambda_{\text{lat}} = 1.0$.

\subsubsection{Evaluation Metrics.} Following the BOP Challenge protocol~\cite{nguyen2025bop}, we report Average Recall (AR) as the primary evaluation metric. AR is computed from three complementary pose-error metrics: Visible Surface Discrepancy (VSD), Maximum Symmetry-Aware Surface Distance (MSSD), and Maximum Symmetry-Aware Projection Distance (MSPD). For each metric, recall is averaged over multiple predefined error thresholds, and the final AR is obtained by averaging the three resulting recall values.

\subsubsection{Datasets.} We evaluate the proposed method on the seven core BOP challenge datasets~\cite{nguyen2025bop}: LM-O\cite{brachmann2014learning}, T-LESS\cite{hodan2017t}, TUD-L\cite{hodan2018bop}, IC-BIN\cite{doumanoglou2016recovering}, ITODD\cite{drost2017introducing}, HB\cite{kaskman2019homebreweddb}, and YCBV\cite{xiang2017posecnn}. Together, these datasets cover a broad range of object categories and challenging scenarios, including severe occlusion, cluttered scenes, and texture-less or symmetric objects. We use default CNOS~\cite{nguyen2023cnos} object detections provided by the BOP\cite{nguyen2025bop}.

\subsection{BOP Benchmark Results}
\label{sec:main_results}

\begin{table*}[t]
\centering
\caption{\textbf{Main results on BOP benchmark datasets.} Comparison of our method against RGB-only coarse pose estimation baselines by reporting the Average Recall (AR) scores on the seven core BOP\cite{nguyen2025bop} datasets.}
\label{tab:main_results}
\resizebox{\textwidth}{!}{%
\begin{tabular}{l c ccccccc c c}
\toprule
\textbf{Method} & \textbf{RANSAC} & \textbf{LM-O} & \textbf{T-LESS} & \textbf{TUD-L} & \textbf{IC-BIN} & \textbf{ITODD} & \textbf{HB} & \textbf{YCB-V} & \textbf{Mean} & \textbf{Time(s)} \\ \midrule
MegaPose~\cite{labbe2022megapose} & \xmark & 22.9 & 17.7 & 25.8 & 15.2 & 10.8 & 25.1 & 28.1 & 20.8 & 15.46 \\
GenFlow~\cite{moon2024genflow} & \xmark & 25.0 & 21.5 & 30.0 & 16.8 & 15.4 & 28.3 & 27.7 & 23.5 & 3.84 \\
GigaPose~\cite{nguyen2024gigapose} & \cmark & 29.9 & 27.3 & 30.2 & 23.1 & 18.8 & 34.8 & 29.0 & 27.6 & \textbf{0.38} \\
FoundPose~\cite{ornek2024foundpose} & \cmark & 39.7 & 33.8 & 46.9 & 23.9 & \underline{20.4} & 50.8 & 45.2 & 37.3 & 1.69 \\
RayPose~\cite{huang2025raypose} & \xmark & 42.1 & 36.9 & 48.3 & 21.8 & -- & -- & 46.2 & 39.1 & -- \\
\textbf{Ours} & \xmark & 45.6 & 24.0 & 46.3 & 35.5 & 19.1 & 63.2 & 54.4 & 41.2 & \underline{0.46} \\
\textbf{Ours-RANSAC} & \cmark & \underline{53.3} & \underline{46.1} & \underline{56.9} & \textbf{61.3} & 18.3 & \underline{67.9} & \underline{61.4} & \underline{52.2} & 0.52 \\
Co-op~\cite{moon2025co} & \cmark & \textbf{59.7} & \textbf{59.2} & \textbf{64.2} & \underline{45.8} & \textbf{39.1} & \textbf{78.1} & \textbf{62.6} & \textbf{58.4} & 0.98 \\ 
\bottomrule
\end{tabular}
}
\end{table*}

Table~\ref{tab:main_results} summarizes the performance of our pipeline against state-of-the-art zero-shot coarse pose estimation methods across the seven core BOP benchmark datasets~\cite{nguyen2025bop}. To evaluate the specific impact of our learned pose solver, we perform a controlled comparison by keeping the correspondence module fixed and evaluating two solver variants: \textbf{Ours}, which uses our proposed neural solver, and \textbf{Ours-RANSAC}, which replaces the PnP-Transformer module with an OpenCV EPnP-RANSAC solver~\cite{fischler1981random, lepetit2009ep}.

Ours achieves a $41.2\%$ mean AR at an inference latency of $0.46$\,s per object, surpassing the prior baselines such as RayPose ($39.1\%$), FoundPose ($37.3\%$), and GigaPose ($27.6\%$), while trailing only Co-op ($58.4\%$). In particular, our neural solver shows strong performance on scenes with clutter and occlusion, reaching $63.2\%$ AR on HB and $35.5\%$ AR on IC-BIN.

When pairing the correspondence module with the EPnP-RANSAC solver, Ours-RANSAC achieves a $52.2\%$ mean AR at an inference latency of $0.52$\,s per object, ranking second overall among zero-shot coarse pose estimation methods. On IC-BIN, it establishes a new state-of-the-art with $61.3\%$ AR, outperforming Co-op ($45.8\%$) by a wide margin, while also achieving strong results on LM-O ($53.3\%$), TUD-L ($56.9\%$), HB ($67.9\%$), and YCB-V ($61.4\%$). 

These results highlight a clear accuracy-for-latency trade-off: EPnP-RANSAC outperforms PnP-Transformer on six of the seven datasets, yielding an $+11.0\%$ boost in mean AR at the cost of an additional $60$\,ms per object. This gap is most pronounced on datasets with high outlier ratios caused by severe symmetries and heavy clutter, such as T-LESS ($24.0\%$ vs.\ $46.1\%$) and IC-BIN ($35.5\%$ vs.\ $61.3\%$), where RANSAC's explicit inlier filtering proves advantageous over global token aggregation. Conversely, on richly textured datasets with lower outlier ratios like LM-O ($45.6\%$ vs.\ $53.3\%$) and YCB-V ($54.4\%$ vs.\ $61.4\%$), this performance gap narrows considerably. A detailed architectural analysis comparing the two solver paradigms is provided in Sec.~\ref{sec:transformer_vs_ransac}.

\begin{figure}[t]
    \centering
    \includegraphics[width=1.0\linewidth]{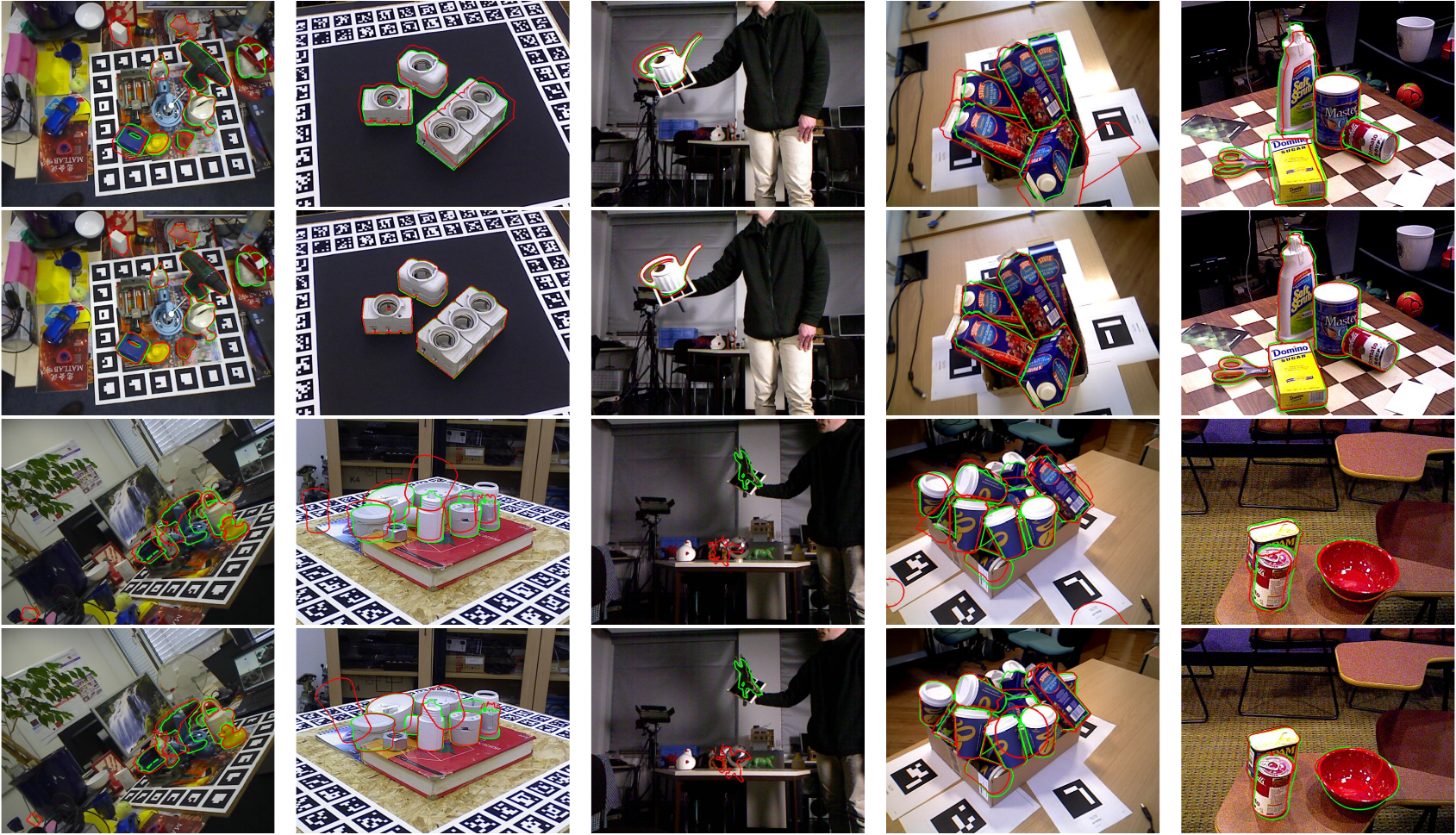}
    \caption{\textbf{Representative pose estimation results on BOP datasets.} Predicted (\textcolor{red}{red}) and ground-truth (\textcolor{green}{green}) object contours overlaid on the input images. Rows \textbf{1} and \textbf{2} represent success cases for PnP-Transformer and PnP-RANSAC, respectively, while Rows \textbf{3} and \textbf{4} show their corresponding failure cases.}
    \label{fig:qualitative_results}
\end{figure}

Figure~\ref{fig:qualitative_results} illustrates qualitative comparisons across diverse BOP benchmark scenarios. On textured objects subject to severe clutter and partial occlusion (Cols.~1, 4, 5), both PnP-Transformer (Row~1) and EPnP-RANSAC (Row~2) achieve precise 3D contour alignment. However, the failure cases (Rows~3–-4) reveal three clear limitations. First, our pipeline performs noticeably worse on textureless objects (Col.~2). Without surface texture, matching relies almost entirely on silhouette edges, which are inherently pose-dependent and offer insufficient signal for reliable 6D pose estimation. Second, symmetric objects (Col.~5, bowl) create multi-modal pose ambiguities that cause errors. Finally, upstream false detections from the CNOS\cite{nguyen2023cnos} object detector (Cols.~3--4) directly corrupt downstream pose predictions. 

\subsection{Ablation Study}

We investigate the contribution of Pose-Encoder to overall pose estimation performance. Since this module is auxiliary, the inference pipeline functions without it by directly feeding the latent vector $\mathbf{z}$ from PnP-Encoder into Pose-Decoder. To assess its impact, we train a baseline for the exact same number of iterations in which Pose-Encoder blocks are removed and the latent alignment loss weight is set to $\lambda_{\text{lat}} = 0.0$. As shown in Table~\ref{tab:ablation}, supervising PnP-Transformer with an additional constraint of aligning correspondence-derived latents with target pose latents yields consistent improvements across the seven core BOP datasets, raising the mean AR from $39.3\%$ to $41.2\%$.

\begin{table}[t]
\centering
\caption{\textbf{Ablation of the Pose-Encoder and latent alignment loss $\mathcal{L}_{\text{lat}}$.} Evaluating the Average Recall (AR) on the seven core BOP datasets with and without latent alignment supervision during training.}
\label{tab:ablation}
\resizebox{\columnwidth}{!}{%
\begin{tabular}{lcccccccc}
\toprule
\textbf{Configuration} & \textbf{LM-O} & \textbf{T-LESS} & \textbf{TUD-L} & \textbf{IC-BIN} & \textbf{ITODD} & \textbf{HB} & \textbf{YCB-V} & \textbf{Mean AR} \\
\midrule
w/o $\mathcal{L}_{\text{lat}}$ (no Pose-Encoder) & 43.0 & 18.4 & 45.9 & 34.6 & 18.7 & 62.1 & 52.1 & 39.3 \\
w/ $\mathcal{L}_{\text{lat}}$ (Full Model) & \textbf{45.6} & \textbf{24.0} & \textbf{46.3} & \textbf{35.5} & \textbf{19.1} & \textbf{63.2} & \textbf{54.4} & \textbf{41.2} \\
\bottomrule
\end{tabular}%
}
\end{table}

\subsection{Emerging Properties of the Learned Latent Space}

We examine the geometric structure and continuity of the learned latent space by extracting target vectors $\bar{\mathbf{z}} \in \mathbb{R}^{256}$ from ground-truth camera poses via Pose-Encoder and projecting them into two dimensions using t-SNE\cite{van2008visualizing}. This visualization evaluates whether the training-time auxiliary branch preserves global and local $\mathrm{SE}(3)$ topological properties when mapping exact rigid transformations. Because the alignment between PnP-Encoder and Pose-Encoder embeddings is explicitly supervised, analyzing Pose-Encoder under clean ground-truth inputs characterizes the ideal manifold topology established by our joint-embedding loss. We emphasize that this analysis reflects the structured target space produced by ground-truth pose inputs; evaluating the inference-time embeddings generated by PnP-Encoder under noisy, predicted correspondence fields is beyond the scope of this evaluation.

\begin{figure*}[t]
    \centering
    \includegraphics[width=\linewidth]{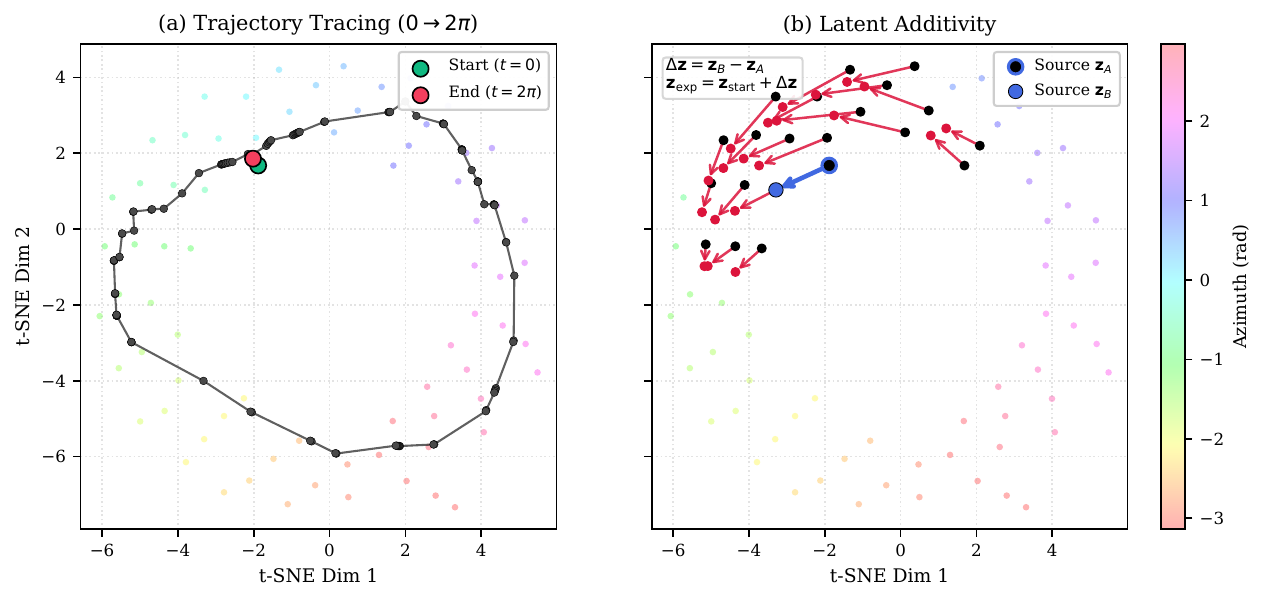}
    \caption{\textbf{Visualization of the learned pose latent space.} 
    (a) \textit{Trajectory Tracing}: Background points represent camera poses sampled across an icosphere, color-coded by azimuth angle. The mapped trajectory of a continuous $360^\circ$ camera orbit around the world Y-axis forms a closed loop in the latent space.
    (b) \textit{Latent Additivity}: A relative displacement vector $\Delta \mathbf{z} = \mathbf{z}_B - \mathbf{z}_A$ extracted from a source pose pair (blue) is applied to neighboring starting points $\mathbf{z}_{\text{start}}$. The resulting transferred shifts (red arrows) demonstrate similar displacements.}
    \label{fig:latent_space_vis}
\end{figure*}

\subsubsection{Global Topology and Trajectory Tracing.} Following our template pose sampling protocol (Sec. \ref{sec:onboarding}), camera viewpoints are sampled across an icosphere and encoded into latent vectors via the Pose-Encoder. As shown in the background distribution of Fig.~\ref{fig:latent_space_vis}(a), when color-coding the t-SNE embeddings by azimuth angle, the projected space naturally organizes into concentric circular rings corresponding to distinct elevation bands, exhibiting continuous color transitions along each ring. To evaluate whether the learned latent space preserves continuous motion paths, we trace a complete camera orbit around the world Y-axis from $0 \to 2\pi$ azimuth rotation in $30^\circ$ increments. As overlaid in Fig.~\ref{fig:latent_space_vis}(a), the sequence of encoded latent vectors forms a closed-loop trajectory that returns to its starting state. This cyclic structure confirms that the Pose-Encoder inherently captures the periodic topology of spatial rotations in $\mathrm{SO}(3)$.

\subsubsection{Latent Space Additivity.} To evaluate whether the learned latent space exhibits the additivity property, we test whether a relative transformation vector can be consistently transferred across different pose locations. We select a reference source pair of camera poses $A, B$ and extract their latent representations $\mathbf{z}_A$ and $\mathbf{z}_B$. The relative motion in latent space is obtained via $\Delta \mathbf{z} = \mathbf{z}_B - \mathbf{z}_A$. We then apply this exact displacement to multiple distinct starting poses $\mathbf{z}_{\text{start}}$, computing predicted arithmetic points $\mathbf{z}_{\text{exp}} = \mathbf{z}_{\text{start}} + \Delta \mathbf{z}$. As visualized in Fig.~\ref{fig:latent_space_vis}(b), applying $\Delta \mathbf{z}$ across different locations produces transferred latent displacements (red arrows) that maintain directionality and magnitude relative to the source shift (blue vector). This confirms that the learned latent space has the additivity property where linear vector arithmetic models camera pose displacements.

\subsection{PnP-Transformer vs. Classical PnP-RANSAC}
\label{sec:transformer_vs_ransac}

As reported in Tab.~\ref{tab:main_results}, replacing our neural pose solver with EPnP-RANSAC\cite{lepetit2009ep} yields a higher AR score across the BOP benchmark datasets\cite{nguyen2025bop}. This performance gap stems from the fundamental algorithmic differences between iterative consensus-based sampling and single-pass global attention aggregation, combined with the decoupled nature of our training pipeline. The correspondence module is trained independently using local matching losses without pose-level supervision, generating correspondence fields without explicit geometric consistency. Given a sufficient ratio of valid correspondences, RANSAC reliably isolates consistent geometric subsets by explicitly partitioning the input field into inliers and outliers based on reprojection error. In contrast, PnP-Transformer processes the entire correspondence set simultaneously in a single forward pass, relying on self-attention mechanisms to summarize geometry into a global latent representation via a class token. While self-attention captures global context effectively, it can remain sensitive to high outlier ratios, which are expectedly higher in textureless and symmetric scenarios, as demonstrated by the performance gaps observed on T-LESS\cite{hodan2017t} and IC-BIN\cite{doumanoglou2016recovering} datasets (Table~\ref{tab:main_results}). Because our current correspondence module utilizes a non-differentiable hard assignment operation ($\operatorname{argmax}$), end-to-end joint optimization could not be applied in this work, leaving the task of resolving geometrically inconsistent correspondences entirely to PnP-Transformer.

Despite the raw accuracy advantage of iterative RANSAC, PnP-Transformer introduces several structural benefits for deep learning pipelines. Unlike stochastic consensus algorithms, PnP-Transformer does not contain discrete sampling operations, making it differentiable. Although the correspondence module and pose solver are trained separately in this work, the introduced pose solver architecture inherently enables gradient flow from downstream pose losses back to feature backbones, opening the door for end-to-end setups where the model adaptively learns to predict only geometrically consistent matches rather than dense, local point fields. Furthermore, while CPU-bound RANSAC implementations exhibit non-deterministic execution times that scale with noise and outlier ratios, PnP-Transformer runs natively on the GPU with constant forward-pass latency, ensuring predictable throughput for real-time applications. Finally, beyond direct pose regression, the model maps arbitrary correspondence fields into a continuous, topologically structured latent space, enabling downstream representation tasks, fast pose indexing, and manifold analysis that are impossible with classical geometric fitting routines.

\section{Conclusion}
\label{sec:conclusion}

We introduced PnP-Transformer, a differentiable neural pose solver that regresses 6D object pose directly from 2D--3D correspondence fields in a single forward pass. By utilizing a joint-embedding architecture, our model structures a continuous latent space that captures the underlying topological geometry of rigid transformations. When integrated into a zero-shot 6D pose estimation pipeline, our framework achieves a $41.2\%$ mean AR score across seven core BOP benchmark datasets at an inference latency of $0.46$\,s/obj, outperforming prior baselines such as RayPose ($39.1\%$), FoundPose ($37.3\%$), and GigaPose ($27.6\%$), while placing behind only Co-op ($58.4\%$). To evaluate the standalone performance of our learned pose solver, we conducted a controlled comparison by replacing PnP-Transformer with an EPnP-RANSAC solver while holding the correspondence module fixed. The EPnP-RANSAC variant achieves a $52.2\%$ mean AR score at the cost of an additional $60$\,ms per object ($0.52$\,s/obj), ranking second overall among zero-shot methods and setting a new state-of-the-art on IC-BIN ($61.3\%$). This comparison reveals a direct accuracy-for-latency trade-off between iterative stochastic sampling and constant-time attention aggregation. Future work includes integrating our differentiable solver into an end-to-end framework to enable joint optimization of correspondence estimation and pose regression. Furthermore, we plan to leverage the learned latent pose representations within Simultaneous Localization and Mapping (SLAM) systems, exploring their utility in Extended Kalman Filtering (EKF) and fast topological pose indexing.


%
%
\bibliographystyle{splncs04}
\bibliography{main}
\end{document}